\pdfoutput=1

\documentclass[11pt]{article}

\usepackage[preprint]{coling}

\usepackage{times}
\usepackage{latexsym}

\usepackage[T1]{fontenc}

\usepackage[utf8]{inputenc}

\usepackage{microtype}

\usepackage{inconsolata}

\usepackage{graphicx}
\usepackage{amsmath}
\usepackage{booktabs, multirow} 
\usepackage{xspace}
\usepackage{color, xcolor,colortbl} 

\newcommand{\model}{VidF4\xspace}
\title{End-to-End Video Question Answering with Frame Scoring Mechanisms and Adaptive Sampling}

\author{
 \textbf{Jianxin Liang\textsuperscript{1}},
 \textbf{Xiaojun Meng\textsuperscript{2}},
 \textbf{Yueqian Wang\textsuperscript{1}},
 \textbf{Chang Liu\textsuperscript{1,3}},
\\
 \textbf{Qun Liu\textsuperscript{2}},
 \textbf{Dongyan Zhao\textsuperscript{1}}
\\
 \textsuperscript{1}Wangxuan Institute of Computer Technology, Peking University
 \\
 \textsuperscript{2}Huawei Noah’s Ark Lab
 \\
 \textsuperscript{3}Center for Data Science, Peking University
 \\
\texttt {\{liangjx,wangyueqian,liuchang97,zhaody\}@pku.edu.cn}\\ 
 \texttt {\{xiaojun.meng, qun.liu\}@huawei.com}
}

\begin{document}
\maketitle

\begin{abstract}

Video Question Answering (VideoQA) has emerged as a challenging frontier in the field of multimedia processing, requiring intricate interactions between visual and textual modalities.
Simply uniformly sampling frames or indiscriminately aggregating frame-level visual features often falls short in capturing the nuanced and relevant contexts of videos to well perform VideoQA.
To mitigate these issues, we propose VidF4, a novel VideoQA framework equipped with tailored frame selection strategy for effective and efficient VideoQA.
We propose three frame-scoring mechanisms that consider both question relevance and inter-frame similarity to evaluate the importance of each frame for a given question on the video.
Furthermore, we design a differentiable adaptive frame sampling mechanism to facilitate end-to-end training for the frame selector and answer generator.
The experimental results across three widely adopted benchmarks demonstrate that our model consistently outperforms existing VideoQA methods, establishing a new SOTA across NExT-QA (+0.3\%), STAR (+0.9\%), and TVQA (+1.0\%). Furthermore, through both quantitative and qualitative analyses, we validate the effectiveness of each design choice. 

\end{abstract}




\section{Introduction}

Video Question Answering (VideoQA) ~\cite{patel2021recent,zhong2022Video} stands at the intersection of computer vision and natural language processing, aiming to enable intelligent systems to understand and respond to questions posed to a given video. A fundamental challenge in VideoQA lies in effectively leveraging both visual and textual modalities to generate accurate answers. Recent advancements in Large Language Models (LLMs) have significantly improved the performance of VideoQA systems. 
However, integrating visual information into LLMs remains a bottleneck, particularly due to the inherent complexity (i.e., temporal dynamics and long-context visual frames) of video.


\begin{figure}[!tb]
    \centering
    \includegraphics[scale=0.4]{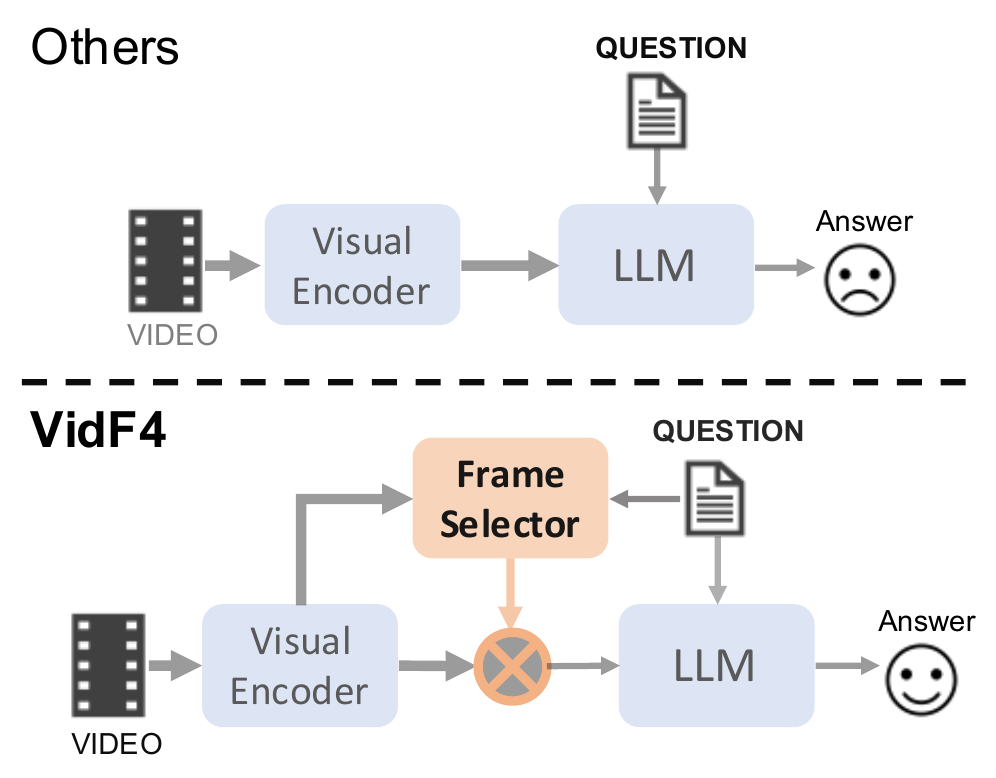}
    \caption{Dataflow of VideoQA models. The width of the arrow represents the amount of video frames.}
    \label{fig:dataflow}
\end{figure}

Existing methods often find and encode the vital features of video data and thus perform jointed learning with textual features from LLMs. 
Simple methods, e.g., uniformly sampling frames or aggregating frame-level features~\cite{pan2023retrieving,ko2023large}, are proposed to integrate these visual features with LLMs. However, they often fail to capture the nuanced temporal relationships and vital contexts of videos, and thus ignore the required visual information to answer the given question.

Addressing this challenge requires the use of intelligent frame selection within VideoQA models~\cite{kim2020dense, kim2021self, yu2023self}. The frame selector serves to identify informative temporal segments of videos that are most relevant to the posed question. By dynamically attending to salient visual cues, a frame selector enables VideoQA models to focus on key moments, thereby improving the accuracy of generated answers.
In previous research, \citet{kim2021self} employs locally aligned attention to effectively focus on the video frames that are relevant to the question. 
SeViLA \citep{yu2023self} attempts a two-stage approach, first training a keyframe localizer using LLMs and then training an answer generator. However, this method may not fully consider the mutual influence and optimization among different stages.

To address these issues, we propose an end-to-end training framework based on frame selection mechanisms, called \model. As shown in Figure~\ref{fig:dataflow}, \model comprises an answer generator consisting of a visual encoder and 
along with a novel frame selector composed of three frame scoring mechanisms and a differentiable adaptive frame sampler.
We particularly emphasize our designed frame selector, which filters out irrelevant frames and prioritizes those containing vital visual context, allowing our model to attend to the most informative frames and thus benefit answering.

Specifically, our frame-scoring mechanisms comprise three components: Question-Frame Similarity (QFS), Question-Frame Matching (QFM), and Inter-Frame Distinctiveness (IFD). They comprehensively evaluate the importance of each frame for a given question. QFS measures the semantic similarity between video frames and the question, QFM evaluates the matching degree between video frames and the question, while IFD considers the redundancy and distinctiveness among video frames. Additionally, to directly optimize the overall VideoQA performance, we employ a differentiable weighted reservoir sampling algorithm ~\cite{efraimidis2006weighted, xie2019reparameterizable} to bridge the frame selector and answer generator. 
Experiments show that this end-to-end training method significantly improves our model's performance.

Overall, extensive experiments demonstrate that our proposed method \model outperforms SOTA models across three popular VideoQA benchmarks: NExT-QA~\citep{xiao2021next} (+0.3\%), STAR~\citep{wu2021star} (+0.9\%), and TVQA~\citep{lei2018tvqa} (+1.0\%). 
We validate the effectiveness of each frame scoring mechanism through both quantitative and qualitative analyses in ablation studies and case studies. Furthermore, we delve into framework design choices and assess resource overheads. In summary, our contributions are three folds:
\begin{itemize}
    \item We design an end-to-end framework to address VideoQA tasks via frame selection. We employ a differentiable weighted sampling method, connecting frame selection with the answer generator. This end-to-end adaptive method optimizes globally, and thus better adapts to the final optimization goal.

    \item We propose three frame scoring mechanisms to comprehensively evaluate the importance of video frames for a given question, providing a basis for models to filter out irrelevant frames and attend to the most informative ones.

    \item We conduct extensive experiments on multiple VideoQA datasets, and results validate the effectiveness of our proposed method, demonstrating superior performance compared to other approaches.

\end{itemize}





\section{Related Work}

Previous VideoQA research has explored various methods to understand and answer questions related to video content. To drive the community forward and comprehensively assess model capabilities, researchers have constructed VideoQA-related datasets using manual or automatic annotation methods\cite{tapaswi2016movieqa, xu2017video,lei2018tvqa,zhao2018open,xiao2021next,wu2021star}, such as NExT-QA \cite{xiao2021next}, STAR \cite{wu2021star}, and TVQA \cite{lei2018tvqa}. These datasets cover multiple aspects including video content, spatiotemporal understanding, and causal reasoning, significantly advancing the field of VideoQA. To better model these features and understand the interaction with other modalities, VideoQA methods have undergone a paradigm shift from early attention mechanisms, memory modules, graph neural networks to pre-trained models\cite{xu2017video, jang2017tgif,khan2020mmft,yang2020bert,yang2021just,lei2021less,wang2022internvideo,ye2023hitea,gao2023mist,wang2023all}. For example, JustAsk\cite{yang2021just} utilizes the HowToVQA69M dataset\cite{yang2021just} for contrastive learning, which includes contrastive learning between a multimodal video-question transformer and an answer transformer. InternVideo\cite{wang2022internvideo} extends a vision transformer pre-trained on images for video representation learning. Recently, with the increasing capabilities of foundation models, some studies have attempted to address VideoQA tasks by leveraging LLMs. For instance, LLaMA-VQA~\cite{ko2023large} introduces multiple auxiliary generation objectives to harness LLaMA's~\citep{touvron2023llama} knowledge of temporal and causal reasoning for understanding the intricate VideoQA task.

Another related research area is keyframe selection. Previous studies on keyframe selection have primarily focused on tasks such as video captioning, video summarization, video classification, and action recognition\cite{hannane2016efficient, huang2019novel,kulhare2016key,singh2021efficient,tang2019fast,yan2020self}. Recently, researchers attempt to incorporate keyframe selection into VideoQA to address the challenges posed by high-dimensional, long-form video data. Two advanced frame-selection-based methods have been proposed. LSTP\cite{wang2024lstp} employs an optical flow-based method to predict multiple segments of video and then utilizes language models through prompts to generate answers. SeViLA \cite{yu2023self} attempts to endow the model with keyframe retrieval capabilities by pretraining from a video moment retrieval/grounding task. It employs LLM multiple times to select keyframes and subsequently generate answers based on these selected keyframes.

Unlike previous methods, our paper proposes three frame scoring mechanisms, which evaluate the importance of each frame for the given question from three perspectives: Question-Frame similarity (QFS), Question-Frame matching (QFM), and Inter-Frame Distinctiveness (IFD). These mechanisms allow the model to filter out irrelevant frames and focus on the most informative ones.
\section{Task Formulation}\label{sec:task}
We target the task of multi-choice video question answering given its popularity. In this task, given a video clip $\mathcal{V}=\{v_1, v_2, ..., v_M\}$
with \(M\) video frames, relevant question \(q\), and \(N\) candidate answers \(\mathcal{A} = (a_1, a_2, ..., a_N)\), the objective for model \texttt{G($\cdot$)} is to select a correct answer \(a^* = \texttt{G(}\mathcal{V}, q,\mathcal{A}\texttt{)}\) from the candidate answer set \(\mathcal{A}\). We formalize this task as a generative task, requiring the model to output the correct answer in a generative manner. For a specific question, leveraging the full video may involve numerous irrelevant or redundant content, which brings too many visual features and could impact the model performance. Therefore, we first introduce a frame selector to eliminate such irrelevant or redundant frames, and thus to identify the most vital visual features for model \texttt{G($\cdot$)} to perform the VideoQA task. In particular, we use the video selector \texttt{F($\cdot$)} to obtain \(\mathcal{V}^* = \texttt{F(}\mathcal{V}, q,\texttt{)}\), where \(\mathcal{V}^*\)=$\{v_{z_1}, v_{z_2}, ..., v_{z_k}\}$ represents an ordered $k$ frames from \(\mathcal{V}\). 


\begin{figure}[!thb]
    \centering
    \includegraphics[scale=0.42]{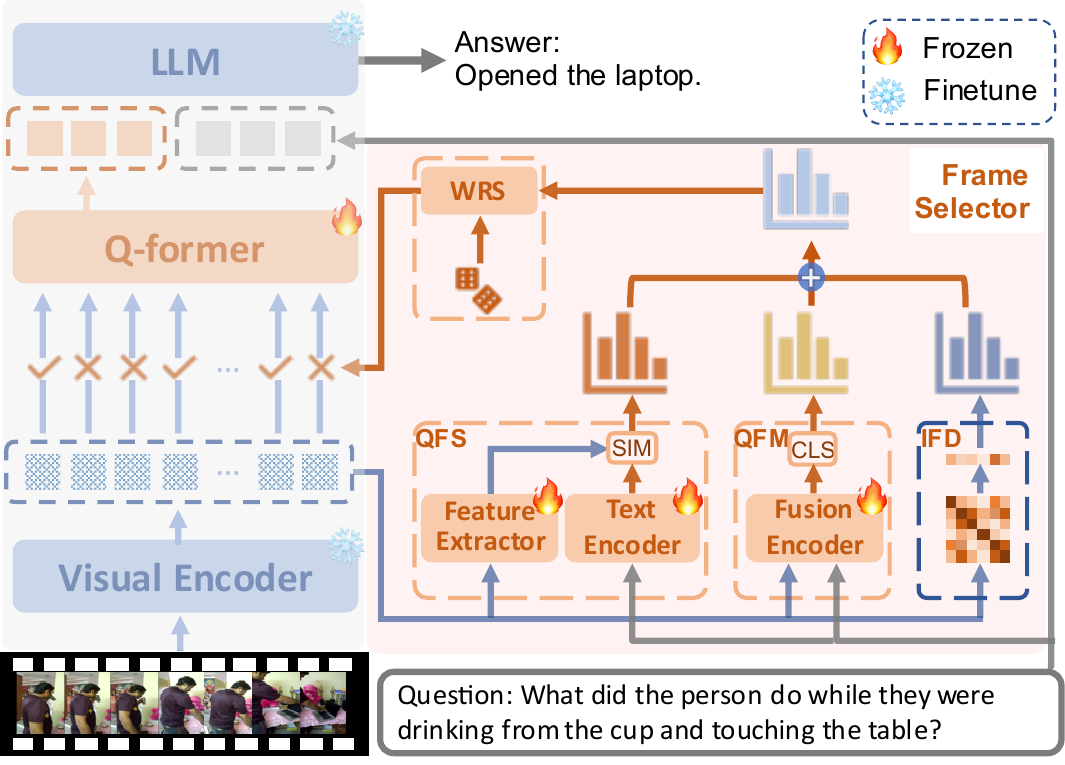}
    \caption{The overall framework, \model, consists of two main components: Answer Generator and Frame Selector. Within the Frame Selector, three scoring mechanisms—QFS, QFM, and IFD—are utilized to evaluate the relevance of frames to the given question. Additionally, there's a differentiable sampler that uses the final score of each frame as the weight for sampling. This adaptive sampler enables \model~to be trained in an end-to-end manner.}
    \label{fig:overall framework}
\end{figure}

Afterwards, the selected video frames \(\mathcal{V}^*\) are passed to an answer generator to generate the final answer. In such an answer generator, a widely used multi-modal perceiver (\textit{i.e.,} Q-former~\cite{li2023blip}) is employed to perform cross-modal attention between the selected visual frames and textual input to achieve the fused representation, often named as visual queries. Finally, we concatenate the visual queries and the textual input that consists of questions, options, and any other descriptions, to form an ultimate prompt as the input for the large language model. This comprehensive process ensures the seamless integration of visual and textual elements, contributing to an enhanced approach to answer generation. The overall modelling process can be expressed as \(a^* = \texttt{G(}\texttt{F(}\mathcal{V}, q,\texttt{)}, q,\mathcal{A}\texttt{)}\). In this paper, we particularly propose a novel method to improve the video selector \(\texttt{F($\cdot$)}\) and study how it affects visual answer generation.



\section{Methodology}


Overall, our method aims to capture the vital visual information from videos that is related to questions, while ensuring significant differences and mutual complementarity among selected video frames to enhance the accuracy of question answering, as illustrated in Figure~\ref{fig:overall framework}. In this paper, we need to address three research questions: (1) \textbf{Question-aware Frame Relevance}: How do we determine the relevance of frames to a given question, ensuring that the selected frames are crucial for answering that question? (2) \textbf{Inter-Frame Distinctiveness}: How do we reduce redundancy among video frames, maintain frames informative while increasing inter-frame distinctiveness, and select those that are less repetitive and mutually reinforcing, thereby providing a more comprehensive visual representation? (3) \textbf{Connecting Frame Selector and Answer Generator}: The connection from the frame selector to the answer generator is often non-differentiable, making existing work~\cite{yu2023self} a two-stage solution to this problem. Therefore, how to backpropagate gradients from the answer generator to the frame selector via sampled and selected frames, thus enabling the model to be trained in an end-to-end manner, and achieve better adaptation to complex VideoQA tasks?

In the following sections, we provide detailed explanations and solutions to these three key questions, outlining our methodology.
\subsection{Question-aware Frame Relevance}\label{sec:selector}

To ensure that the selected video frames \(\mathcal{V}^*\) are crucial for answering specific questions, we introduce question-aware frame scoring mechanisms to quantify the importance of video frames given an input question \(q\). These proposed mechanisms finely assess the relevance of each frame to this given question and assign an importance score to each frame. Our method operates on the principle that the score of a frame is positively correlated with its relevance to the question \(q\); the higher the score, the stronger the correlation is considered to be between this frame and the question. We accomplish this task by leveraging two mechanisms: similarity and matching scores between frames and questions to obtain the selected video frames \(\mathcal{V}^*\).

In particular, as shown in Figure~\ref{fig:overall framework}, the first one is \textbf{Q}uestion-\textbf{F}rame \textbf{S}imilarity ($\textbf{QFS}$). We first obtain the representation $\textbf{\textit{h}}_i^v$ of frame $v_i$ through the visual encoder \texttt{E$_v$($\cdot$)}. Then, we continue to utilize the feature extractor \texttt{E$_e$($\cdot$)} to further extract visual features $\textbf{\textit{h}}_i^e$ most relevant to the text. We also use the text encoder \texttt{E$_t$($\cdot$)} to obtain the text representation $\textbf{\textit{h}}^q$. We use the cosine similarity between the representations $h^q$ and $h_i^e$ as the score for QFS:
\begin{equation}
    h_i^e = \texttt{E}_e \texttt{(} h_i^v \texttt{)}, h^q = \texttt{E}_q \texttt{(} q \texttt{)},
\end{equation}
\begin{equation}\label{eq:dual}
    \mathcal{S}_{QFS} \texttt{(} v_i, q \texttt{)} = \texttt{cos} \texttt{(} W_e \cdot h_i^e, W_q \cdot h^q \texttt{)},
\end{equation}
where $W_{e}$ and $W_{q}$ are linear transformations that map $h_i^{e}$ and $h^{q}$ to lower-dimensional representations.
The second one is \textbf{Q}uestion-\textbf{F}rame \textbf{M}atching (\textbf{QFM}), where we use the multimodal fusion encoder \texttt{E$_f$($\cdot$)}'s output embedding of the \texttt{[CLS]} token as the joint representation of the frame-question pair, and then feed it into a multi-layer perception
to obtain the final matching score:
\begin{equation}\label{eq:dual}
    \mathcal{S}_{QFM} \texttt{(} v_i, q \texttt{)} = \sigma \texttt{(} W_f \cdot \texttt{E}_f \texttt{(} h_{i}^v, q \texttt{)} \texttt{[CLS]} \texttt{)}.
\end{equation}
where $W_f$ represents trainable parameters, and $\sigma \texttt{(} \cdot \texttt{)}$ is
the \texttt{sigmoid} function. 




\subsection{Inter-Frame Distinctiveness}\label{sec:ifd}
When using the question-aware frame scoring mechanisms, a potential issue arises: the selected frames may exhibit high similarity, \textit{i.e.,} contain very similar content (refer to Figure~\ref{fig:example}, Q2 in appendix), leading to information redundancy and thus insufficient coverage in terms of the other vital content since the total selected frames are limited. 
Certainly, we can try to select as many frames as possible, but it shows a less efficient performance with obviously large inference cost in our experiments.

Therefore, to address this concern, we introduce an \textbf{I}nter-\textbf{F}rame \textbf{D}istinctiveness (\textbf{IFD}) scoring mechanism to assess the distinctiveness between each frame and all others. Higher scores indicate lower similarity to other frames, suggesting greater distinctiveness. Specifically, as illustrated in Figure~\ref{fig:overall framework}, the distinctiveness $\mathcal{D}$\texttt{(}$v_i,v_j$\texttt{)} between the $i$-th and $j$-th frames is defined as:
\begin{equation}\label{eq:dual}
    \mathcal{S} \texttt{(} v_i, v_j \texttt{)}= cos\texttt{(} \texttt{norm(} h_{i}^v\texttt{)}, \texttt{norm(}  h_{j}^v\texttt{))},
\end{equation}
\begin{equation}\label{eq:dual}
    \mathcal{D}\texttt{(}v_i, v_j\texttt{)}= 1-\mathcal{S}\texttt{(}v_i, v_j\texttt{)},
\end{equation}
where $\mathcal{S}\texttt{(}v_i, v_j\texttt{)}$ is the similarity between the normalized representations of the $i$-th and $j$-th frames. Further, the distinctiveness of the $i$-th frame is determined by the average discrimination from all other frames:
\begin{equation}\label{eq:dual}
    \mathcal{S}_{IFD}\texttt{(}v_i\texttt{)}= \frac{\sum_{j\neq i}^M \mathcal{D}\texttt{(}v_i,v_j\texttt{)} }{M-1}.
\end{equation}
We conducted extensive experiments, including ablation study in Table~\ref{tb:ablation} and case study illustrated in Figure~\ref{fig:example}, to demonstrate the effectiveness of IFD. Results show that this function effectively alleviates the issue of excessive redundancy in the selected frames. Refer to Section~\ref{sec:ablation} for experimental details.


Finally, the overall score for the $i$-th frame is given by $\mathcal{S}_{v_i}$ which comprehensively considers the similarity and matching degree between this frame and the given question \(q\), and inter-frame distinctiveness among all frames:
\begin{equation}\label{eq:dual}
    \mathcal{S}_{v_i,q} = \mathcal{S}_{QFS}\texttt{(}v_i,q\texttt{)} + \mathcal{S}_{QFM}\texttt{(}v_i,q\texttt{)} + \mathcal{S}_{IFD}\texttt{(}v_i\texttt{)},
\end{equation}

By now, after targeting the first two mentioned research questions, our proposed integrated score serves as a comprehensive evaluation metric for the question-aware frame selector. Based on this, we are able to assign scores to all frames and obtain the vital video frames \(\mathcal{V}^*\).
However, in terms of the third research question, there still exists a gap between the scoring mechanisms and the final answer generator.
To bridge this connection, we employ a differentiable algorithm (refer to Section~\ref{sec:topk}) to select frames as input for subsequent question answering, thus making our whole framework as an end-to-end trainable solution.

\subsection{Frame Sampling with Scores}\label{sec:topk}





Given $\mathcal{S}_{v_i,q}$ from three scoring mechanisms for each video frame, 
instead of consistently selecting the same frames solely based on the score magnitudes,
we sample video frames with these scores as weights to obtain the final selected set \(V^*\).
In particular, we employ the Weighted Reservoir Sampling (WRS)~\cite{efraimidis2006weighted}, a method where the probability of sampling a frame is directly proportional to its given weight.
This adaptive sampling method introduces subtle variations in the video frames sampled into each training iteration, which gives our model a strong capability to handle various cases, even if our scoring mechanisms are not perfect yet.

Additionally, we aim to train our model in an end-to-end manner without introducing any auxiliary losses for individual modules. Due to the non-differentiable nature of the sampling operation, to enable the use of backpropagation for updating the parameters of the scoring mechanism, we employ a reparameterization technique \texttt{RelaxedTopK($\cdot$)}~\cite{xie2019reparameterizable} which introduces Gumbel Softmax on top of WRS, enabling differentiable subset sampling. \texttt{RelaxedTopK($\cdot$)} takes all frame scores \(\mathcal{S}= \{\mathcal{S}_{v_1,q}, \mathcal{S}_{v_2,q},..., \mathcal{S}_{v_M,q}\}\), sampled number $k$, and a temperature parameter $\tau > 0$ as input. 
It returns a set of relaxed $k$-hot frames, where the probability $p\texttt{(}v_i\texttt{)}$ of selecting frame $v_i$ is calculated using the relaxed distribution of $\mathcal{S}$ with the temperature parameter $\tau$:
\begin{equation}
     p\texttt{(}v_i\texttt{)} = \frac{\exp\texttt{(}\mathcal{S}_{v_i,q}/\tau\texttt{)}}{\sum_j\exp\texttt{(}\mathcal{S}_{v_j,q}/\tau\texttt{)}}.
\end{equation}
As $\tau \rightarrow 0$, the results returned by \texttt{RelaxedTopK($\cdot$)} tend to converge towards the frames corresponding to the top $k$ values in $\mathcal{S}$.
Therefore, the final sampled frames set is: 
\begin{equation}
    \begin{split}
        \mathcal{V}^* &= \texttt{RelaxedTopK}\texttt{(} \mathcal{V},\mathcal{S},k,\tau\texttt{)} \\
    &= \{v_{z_1}, v_{z_2}, ..., v_{z_k}\},
    \end{split}
\end{equation}
where $1 \leq z_1 < z_2 < ... < z_k \leq M$ are the indices preserving the temporal order.
Note that in the training, frames are sampled using this adaptive~\texttt{RelaxedTopK($\cdot$)}, while in the inference, frames are selected from the top-K highest scores in $\mathcal{S}$ ($\tau \rightarrow 0$).

Finally, we feed the selected video frames $\mathcal{V}^*$ to the answer generator in the trainable fashion to bridge the gap mentioned in our third research question. Note that our training objective is only to optimize the softmax cross-entropy loss between the predicted similarity scores of candidate answers and ground truth, as also discussed in Section \ref{sec:task}.


\begin{table*}[hbt]
\centering
\resizebox{\textwidth}{!}{
\begin{tabular}{c|c|cccc|ccccc|c}
\toprule
\multirow{2}{*}{Model}  &\multirow{2}{*}{\#Frames} &\multicolumn{4}{c|}{NExT-QA}  &\multicolumn{5}{c|}{STAR}  &\multirow{2}{*}{TVQA}   
\\
&  & Tem. & Cau. & Des. & Tot. & Int. & Seq. & Pre. & Fea. & Tot. &   \\ 
 
 \midrule
 \multicolumn{10}{l}{\textit{Non-LLM Models}}\\


Just Ask &20  &51.4  &49.6  &63.1  &52.3  &-   &-   &-   &-     &-  &-\\

All-in-One & 32   &48.6  &48.0  &63.2  &50.6  &47.5  &50.8  &47.7  &44.0  &48.9  &-   \\

MIST &32  &56.6  &54.6  &66.9  &57.1  &55.5   &54.2   &54.2   &44.4   &54.0  &-    \\

HiTeA & 16  &58.3  &62.4  &75.6  &63.1  &-  &-  &-  &-  &-    &- \\
InternVideo & 8   &58.5  &62.5  &75.8  &63.2  &62.7  &65.6  &54.9  &51.9  &58.7  &57.2    \\
\midrule

\multicolumn{10}{l}{\textit{LLM-based Models}} \\
BLIP-2$^{voting*}$ &  4  &65.2  &70.1  &80.1  &70.1  &52.3  &54.8  &49.0  &51.2  &51.8 &54.5   \\
BLIP-2$^{concat*}$ & 4    &68.1  &72.9  &81.2  &72.6  &65.4  &69.0  &59.7  &54.2  &62.0 &59.8   \\

LLaMA-VQA & 10 &\underline{69.2}  &72.7  &75.8 &72.0  
&66.2  &67.9  &57.2  &52.7  &65.4  &-   \\

LSTP & 4  &66.5  &72.8  &81.2  &72.1  &-  &-  &-  &-  &-    &- \\

SeViLA$^{\dag}$ & 4   &67.7  &72.1  &\underline{82.2} &73.4  &63.1  &\underline{70.3}  &\textbf{63.4}  &\underline{56.1}  &66.3  & 60.6  \\

SeViLA$^{\dag}$ & 8   &67.0  &\underline{73.8}  &81.8 &\underline{73.8} 
&\underline{66.4}  &70.3  &\underline{61.2}  &55.7  &\underline{67.2} &\underline{60.8}  \\

\midrule
\model & 8   & \textbf{69.6} & \textbf{74.2} & \textbf{83.3} & \textbf{74.1} & \textbf{68.4} & \textbf{70.4} & 60.9 & \textbf{59.4} & \textbf{68.1} & \textbf{61.8}

\\ \bottomrule

\end{tabular}
}
\caption{Model comparison on three benchmarks. Tem., Cau., Des., Tot. denote to Temporal, Causal, Description, Total accuracy respectively. Int., Seq., Pre., Fea. denote to Interaction, Sequence, Prediction, Feasibility respectively. \dag~denotes to our trained SeViLA model by using its official open-sourced codes.
We \textbf{bold} the best numbers, and \underline{underlined} the second-best numbers.}
\label{tb:main results}
\end{table*}





\section{Experimental}
\subsection{Setup}
\paragraph{Datasets.}

We conduct extensive experiments on three popular VideoQA datasets: NExT-QA \cite{xiao2021next}, STAR \cite{wu2021star} and TVQA \cite{lei2018tvqa}, which demand causal and temporal reasoning abilities. We employ the most used standard answer accuracy as the evaluation metric for all datasets. 
\paragraph{Baselines.} We compare our method with two types of baselines: non-LLM and LLM-based models. For non-LLM methods, we use recent SOTA models, including Just Ask \cite{yang2021just}, All-in-One \cite{wang2023all} and MIST \cite{gao2023mist}, HiTeA \cite{ye2023hitea} and InternVideo \cite{wang2022internvideo}. For LLM-based models, we use SOTA models such as BLIP-2 \cite{li2023blip}, LLaMA-VQA \cite{ko2023large}, LSTP \cite{wang2024lstp} and SeViLA \cite{yu2023self}. Among baselines, LSTP and SeViLA are advanced models that also employ frame selection in VideoQA. 

See \textbf{Appendix~\ref{app:setup}} for implementation details.

\subsection{Overall Performance}
We provide the overall evaluation results of our method and baselines in Table~\ref{tb:main results}. 
The comparison between the top and middle groups highlights the significant advantage of using LLM. Even without extensive pretraining on video-text corpora, methods employing LLM consistently outperform those without LLM. This underscores the effectiveness of leveraging LLMs for VideoQA.
\model, aided by three frame scoring mechanisms and frame sampling strategies, comprehensively outperforms the middle group methods, 
It surpasses SeViLa, which employs LLM for frame selection, and even outperforms LLaMA-VQA using 7B parameters.
This showcases \model's clear superiority over previous SOTA approaches.

Specifically, \model excels on NExT-QA, surpassing other SOTAs across all question types. Particularly noteworthy is the substantial improvement in the 
\textit{Description} type, where \model outperforms the current SOTA by 1.1\% (83.3 vs. 82.2), contributing to an overall accuracy increase of over 0.3\%. 
Similarly, \model exhibits superior performance on STAR, surpassing the SOTA model SeViLA by 0.9\%. Notably, in the \textit{Interaction} type, \model outperforms LLaMA-VQA by 2.2\% (68.4 vs. 66.2) and SeViLA by 2.0\% (68.4 vs. 66.4). This pattern is consistent with the performance observed on the NExT-QA, with significant improvements in the \textit{Interaction} and \textit{Description} type. After checking examples from these datasets, we find that the \textit{Interaction} and \textit{Description} types are more specific questions than others, which often require a strong correlation with video frames to perform well. 
Similarly, in the \textit{Feasibility} type, \model outperforms SeViLA by 3.3\% (59.4 vs. 56.1). \textit{Feasibility} type questions, such as '\textit{Q4: What else is the person able to do with the bag?}' in Figure~\ref{fig:example}, require the model to understand the object's current and past states to predict its next possible state change. 
It further reveals the effectiveness of our proposed scoring mechanisms.


\subsection{Ablation Study}\label{sec:ablation}
To investigate the effect of each component in our framework, we conduct an extensive ablation study in Table~\ref{tb:ablation}. The \textit{Random} and \textit{Uniform} respectively refer to selecting 8 frames randomly or uniformly from video frames, used as input for the answer generator.


The \textit{Random} and \textit{Uniform}, employed as comparative baselines, exhibit superior performance compared to all Non-LLM models in Table~\ref{tb:main results}, which thus strongly motivates our work on designing an intelligent model to handle complexities in selecting proper video frames, such as frame relevance and redundancy. This observation again highlights the efficacy of LLM in VideoQA. 
Furthermore, We note that the performance of \textit{Random} on the \textit{Interaction} is 63.5\% (VidF4 68.4\%), while on \textit{Feasibility}, \textit{Random} frame selection achieves 56.9\%, even surpassing SeViLA's 55.7\%, whereas \model achieves 59.4\%. This underscores the importance of our frame selection mechanism in capturing the necessary correlations within the video to significantly enhance task performance. 

\begin{table}[!tb]
\begin{tabular}{c|ccccc}
\toprule
\multirow{2}{*}{Setting}  &
\multicolumn{5}{c}{\textbf{STAR}} \\
&\textbf{Int.}  &\textbf{Seq.}  &\textbf{Pre.}  &\textbf{Fea.}  &\textbf{Tot.}  \\ 
 
 \midrule
 \midrule
 
 \multicolumn{1}{c|}{Random }  &63.5  &68.0  &59.1  &56.9  &64.9
 \\
 \multicolumn{1}{c|}{Uniform } &64.4  &69.0  &58.2  &55.3  &65.6 \\
 \midrule

\rowcolor{gray!30}\model  &68.4  &70.4  &60.9  &59.4  &68.1  \\

\rowcolor{gray!20}w/o QFS  &67.6  &69.8  &60.3  &59.6  &67.5   \\

\rowcolor{gray!15}w/o QFM  &66.6  &70.1  &59.6  &58.6  &67.2   \\

\rowcolor{gray!10}w/o IFD &68.3  &69.8  &59.4  &56.3  &67.4   \\


\midrule

\end{tabular}
\caption{Ablation study for each component of \model. We perform each experiment using pairwise combinations of three components on STAR.}
\label{tb:ablation}
%
\end{table}

 







From the last three rows of the results, it is evident that excluding any of the three scoring mechanisms leads to a decline in \model's performance. In particular, the omission of QFS results in a decrease of 0.8\% in the \textit{Interaction} category, with Sequence and Prediction experiencing drops of 0.6\% each. 
Substantial declines in \textit{Prediction} and \textit{Feasibility} by 1.5\% and 3.1\%, respectively, occur when IFD is removed. 
It highlights that using IFD to remove redundancy thus improving unique content coverage, plays an important role in \textit{Prediction} and \textit{Feasibility} type.
Not using QFM leads to a significant 1.8\% drop in the \textit{Interaction} category. This indicates that QFM has a substantial positive impact on performance in the \textit{Interaction} category, particularly when dealing with question-frame matching.
%

Overall, these results in the ablation study emphasize the interplay of our three scoring mechanisms in \model and their impact on performance across different question types. The combined effect of these components proves to be the key factor contributing to \model's outstanding performance. 

\subsection{Discussion}




\paragraph{\textbf{The choice of the number of selected frames $k$ in training.}} During experiments, we systematically vary the number of selected frames used in training, exploring a range of frame quantities to assess their impact on performance. We use the same number of frames for evaluation as well as for training. The result is shown in Table~\ref{tb: numframes}. We observe that while increasing the selected frames generally leads to performance improvement, there is a diminishing returns trend. After selecting $k$=8 frames, the overall accuracy of \model begins to plateau or decrease, indicating that the information provided by video frames might have reached a saturation point.

By observing the changes of model performance across internal question types in Table~\ref{tb: numframes}, we find that, with an increase in the number of frames, the model performance on \textit{Int.} and \textit{Fea.} type falls into a consistent pattern of an initial increase followed by a decrease. Performance reaches a peak and then exhibits a downturn, suggesting that additional frames might not necessarily bring stable performance improvement and could even potentially introduce noise and errors, leading to a performance decline. Conversely, the performance on \textit{Seq.} and \textit{Pre.} type consistently increases with using more frames. This suggests an understandable claim that, for sequences and problems requiring prediction, a larger number of frames can assist the model in better modeling the video context. 
\begin{table}[!tb]
\begin{tabular}{c|ccccc}
\toprule
\multicolumn{1}{c|}{\#Frames} & 
\multicolumn{5}{c}{\textbf{STAR}} \\
$k$&\textbf{Int.}  &\textbf{Seq.}  &\textbf{Pre.}  &\textbf{Fea.}  &\textbf{Tot.}  \\ 
 \midrule
 
 \midrule
\rowcolor{gray!5}32$\xrightarrow{}$4 &62.3  &64.7  &57.4  &54.9  &62.6    \\
\rowcolor{gray!30} 32$\xrightarrow{}$8 &68.4  &70.4  &60.9  &59.4  &68.1    \\
\rowcolor{gray!20}32$\xrightarrow{}$12 &67.2  &70.6  &62.3  &61.0  &68.0   \\
\rowcolor{gray!10}32$\xrightarrow{}$16 &65.3  &71.5  &63.0  &59.4  &67.8    \\

\midrule

\end{tabular}
\caption{Comparison of model performance across different \( k \) frames (\( k = \{4, 8, 12, 16\} \)) in training. We use the same number of frames for evaluation as well as for training.
}
\label{tb: numframes}
%
\end{table}

In conclusion, the optimal \( k \) varies for different types of questions. Achieving optimal overall performance requires a delicate balance between the number of selected frames and the distribution of data across different question types. Our paper suggests that future research in video QA should delve deeper into each question type and explore how individual efforts truly enhance performance.



\begin{figure}[!tb]
    \centering
    \includegraphics[scale=0.45]{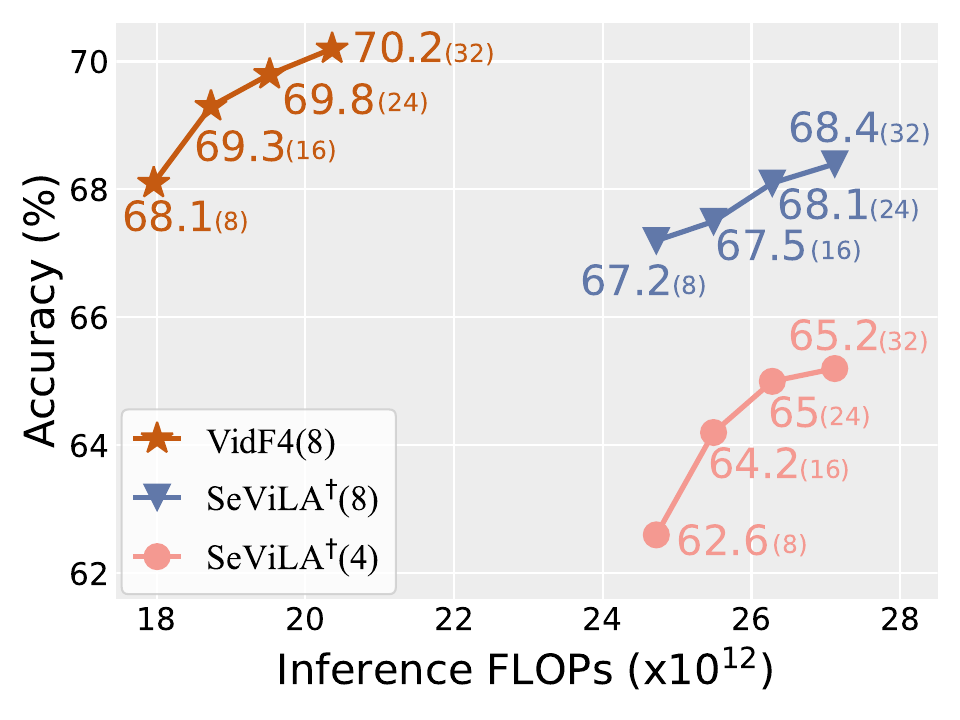}
    \caption{Model performance and FLOPs comparison on STAR using different numbers of frames during inference, where the same color represents the same model checkpoint. For example, the brown stars denote VidF4(8), which corresponds to the model trained at \( k = 8 \), and evaluated under various $k'$ during inference (\( k'=\{8,16,24,32\} \)).
    }
    \label{fig:topK_frame}
\end{figure}

\paragraph{\textbf{The choice of the number of selected frames $k'$ in inference.}}
Previous work \cite{yu2023self,wang2024lstp} typically defaults to using the same number of frames for testing model performance as during training. We denote \( k' \) as the number of frames used during inference, which typically equals \( k \) (\( k' = k \)). Furthermore, we compare the performance and efficiency of models with varying numbers of frames $k'$ during inference, where \( k' \) might not equal \( k \).
The results in Figure~\ref{fig:topK_frame} show the performance of VidF4(8), SeViLA$^\dag$(8) and SeViLA$^\dag$(4) when tested under different numbers of frames ($k'$=\{8, 16, 24, 32\}).



\begin{table}[!tb]
\resizebox{\linewidth}{!}{
\begin{tabular}{c|cc|c|c}
\toprule
\multirow{2}{*}{Models} & 
  \multicolumn{2}{c|}{\textbf{ Generator}} &\multirow{1}{*}{\textbf{Frame}} &\multirow{2}{*}{\textbf{Other}} \\
 &\textbf{Vis. Enc.} &\textbf{LLM} &  \textbf{Selector} &  \\ 
 \midrule
 

SeViLA &\multirow{2}{*}{16.66} &\multirow{2}{*}{0.83}   &7.35    &0.24     \\
VidF4 &   &  &0.44 &     0.05\\

\midrule

\end{tabular}
}
\caption{The average FLOPs ($\times10^{12}$) per instance for each component of the model during inference, e.g., for an example of a video-question pair with 32 frames in the video and 15 tokens in the question, the FLOPs for each component are as follows: QFS is 192.94 x10$^9$, QFM is 280.43 x10$^9$, IFD is 4.19 x10$^6$, totaling 0.47 x10$^{12}$ FLOPs.}
\label{tb: flops}
%
\end{table}

It can be observed that the model's inference performance generally improves with the increase in the number of frames used. Specifically, VidF4(8)'s performance continues to improve, with a maximum improvement of 2.1\% (68.1 vs. 70.2), albeit with an additional 13\% computational cost.
Interestingly, further analysis reveals that models trained with $k$=8 frames and tested with $k'$ frames outperformed models directly trained with $k=k'$ frames in inference, as depicted in Table~\ref{tb: numframes}. 
Specifically, in Figure~\ref{fig:topK_frame}, the performance of VidF4(8) using \( k' = 16 \) frames for inference (69.3) surpasses that of VidF4(16) using \( k' = 16 \) frames for inference (67.8) as shown in Table~\ref{tb: numframes}. Additionally, VidF4(8) achieves approximately 30\% memory savings compared to VidF4(16) during training.
Our \model achieves consistent improvement with the inclusion of more frames during both training and inference. This suggests that \model effectively leverages additional visual context to enhance its understanding and answering capability, especially compared to other baselines. Our method \model can select a different number of frames for inference based on computational and memory constraints.

Overall, \model performs better than SeViLA, positioned closer to the top-left corner of the graph. Specifically, VidF4(8) outperforms SeViLA$^\dag$(8) by 0.9\% to 1.8\% in terms of performance, while reducing the computational cost during inference by 25\% to 28\% in FLOPs. To visually compare the computational overhead of each component, we conduct a breakdown of the computational cost during model inference, as shown in Table~\ref{tb: flops}. The reduction in the computational cost of \model compared to SeViLA mainly stems from the frame selector. SeViLA employs LLM to score each frame individually and selects the frame with the highest score, while \model utilizes a simpler, lower-parameter, and lower-computation scoring mechanism, accounting for only about 2\% of the computational cost.


\paragraph{Applied with other MLLMs.}
We conduct experiments to demonstrate that our method can be applied to other MLLMs: Video-LLaMA (LLaMA, 7B) \cite{zhang2023video} and Blip2-Flan-T5 (T5, 11B) in Table~\ref{tb:otherllm}. We can observe that after incorporating \model, the overall performance of these models has improved. VideoLLaMA improved by 1.2\% (66.5 v.s. 67.7), and Blip2-Flan-T5-xxl improved by 1.4\% (68.4 v.s. 69.8). 
The results highlight the robustness and generalization of our frame selection strategy.
\begin{table}[!hb]
\renewcommand{\thetable}{5}
\resizebox{\linewidth}{!}{
\begin{tabular}{c|ccccl}
\toprule
\multirow{2}{*}{Setting} & 
\multicolumn{5}{c}{\textbf{STAR}} \\
&\textbf{Int.}  &\textbf{Seq.}  &\textbf{Pre.}  &\textbf{Fea.}  &\textbf{Tot.}  \\ 
 \midrule
 
Video-LLaMA&64.6  &68.5  &62.0  &66.5  &66.5\\

\rowcolor{gray!18} w. \model (our) &66.8  &69.2  &63.8  &66.8   &67.7\textcolor[rgb]{0,0.7,0.33}{(+1.2)}   \\

\midrule
Blip2-Flan-T5-xxl  &69.9  &70.5  &60.4  &57.1  &68.4\\

\rowcolor{gray!18} w. \model (our) &70.9  &71.8  &63.5  &58.0  &69.8\textcolor[rgb]{0,0.7,0.33}{(+1.4)}   \\

\end{tabular}
}
\caption{Combining \model with Other MLLMs}

\label{tb:otherllm}
\vspace{-7mm}
\end{table}

\section{Conclusion}

In conclusion, we present a novel approach for video question answering, incorporating three frame selection mechanisms and leveraging a large language model for answer generation. We introduce a differentiable relaxed top-K sampling algorithm to enable end-to-end training, ensuring seamless gradient flow between the answer generator and frame selector. This promotes the tight integration of visual and textual information, allowing our model to identify key moments in videos and then generate more accurate answers. Extensive experiments verify the effectiveness and competitive performance of our method.

\appendix

\section{Experiments Setup}
\label{app:setup}
\subsection{Dataset Details}
\label{app:data}

\textbf{NExT-QA} is a VideoQA benchmark targeting the explanation of video content. The video in NExT-QA primarily encompasses aspects of daily life, social interactions, and outdoor activities, featuring three types of questions: \textit{Temporal} (Tem), \textit{Causal} (Cau), and \textit{Descriptive} (Des). It contains 5.4k videos and about 52K manually annotated question-answer pairs, each QA pair comprises one question and five candidate answers.

\textbf{STAR} is oriented towards real-world reasoning scenarios, encompassing four question types, namely \textit{Interaction} (Int), \textit{Sequence} (Seq), \textit{Prediction} (Pre), and \textit{Feasibility} (Fea). STAR contains 22K Situation Video Clips and 60K Situated Questions.

\textbf{TVQA} is a multi-choice QA benchmark containing 152k QA pairs and 21k video clips from 460 hours of video. It features videos from genres such as sitcoms, medical dramas, and crime series. 


\begin{table}[h]
\centering
\resizebox{\linewidth}{!}{
\begin{tabular}{c|ccc}
\toprule
\multirow{1}{*}{} & 
\multicolumn{1}{c}{\textbf{NExT-QA}}  &\multicolumn{1}{c}{\textbf{STAR}}  &\multicolumn{1}{c}{\textbf{TVQA}} \\

 \midrule
 
\rowcolor{white!10}\#videos &5.4k  &22k  &21k      \\
\rowcolor{white!20}\#questions &52k  &60k  &152k    \\

\midrule
\end{tabular}
}
\caption{Statistics of the datasets we used.}
\label{tab:statistics}
\end{table}



\subsection{Baselines Details}
\label{app:baseline}

\textbf{BLIP-2} processes all frames by voting or concatenating them and then uses LLM to generate the final answer.

\textbf{LLaMA-VQA} is built on the basis of LLaMA-7B \cite{touvron2023llama}, enabling the model to understand the complex relationships between videos, questions, and answers by constructing multiple auxiliary tasks.

\textbf{LSTP} adopts the BLIP-2 architecture and uses optical flow for frame selection, followed by using LLM to generate answers. 

\textbf{SeViLA} relies on a multi-stage training process. During the inference, it first utilizes BLIP-2 and LLMs for frame selection and then uses BLIP-2 and LLM again for answer generation. 

It's worth noting that, according to the original paper\cite{yu2023self}, SeViLA's performance is reported to be better with 4 frames compared to 8 frames, which contrasts with our findings. We pretrain and fine-tune SeViLA with different numbers of frames using official code and hyperparameters and found that the performance was better with 8 frames compared to 4 frames.

\section{Implementation Details}
\label{app: implementation}
We use BLIP-2 as our answer generator. Specifically, we employ ViT-G \cite{fang2023eva} as the visual encoder and initialize FlanT5-XL \cite{chung2022scaling} (3B parameters) as the LLM. We use the Qformer trained in the \textbf{second-stage} of BLIP-2 to connect ViT and FlanT5-XL, which is trained on a large dataset of image-text pairs to align visual representations to the representation space of the LLM.
For the feature extractor \texttt{E$_e$($\cdot$)} and text encoder \texttt{E$_t$($\cdot$)} in QFS, as well as the multimodal fusion encoder \texttt{E$_f$($\cdot$)} in QFM, we employ the Transformer encoder architecture. \texttt{E$_e$($\cdot$)}, \texttt{E$_t$($\cdot$)} and \texttt{E$_f$($\cdot$)} share parameters (188M parameters), and they are all initialized with the parameters of Qformer pretrained in the \textbf{first-stage} of BLIP-2. 

We denote the model trained with \( k \) frames as VidF4(\( k \)). Unless otherwise specified, the model uses the same number of frames \( k \) for both testing and training, consistent with previous research. We use the frame selector to select $k=8$ frames from a total of 32 frames in the video, and $\tau=0.1$ for WRS. We conducted experiments using 4 $\times$ A100 (80G) GPUs. For each dataset, the training batch size per GPU was set to 8, with a learning rate of 3e-5. We implemented a warm-up of 1000 steps and trained the model for 10 epochs.

\section{More Ablation}
We conduct the
ablation experiments on two additional datasets in Table~\ref{app:ablation}.

\begin{table}[!hb]
\vspace{-2mm}
\renewcommand{\thetable}{6}
\begin{tabular}{c|cccc|c}
\toprule
\multirow{2}{*}{Setting}  &
\multicolumn{4}{c|}{\textbf{NExT-QA}}& \multicolumn{1}{c}{\textbf{TVQA}} \\
&\textbf{Tem.}  &\textbf{Cau.}  &\textbf{Des.}  &\textbf{Tot.}  &\textbf{Acc.}\\ 
 
 \midrule
 \midrule
 

\rowcolor{gray!30}\model  &69.6  &74.2  &83.3  &74.1  &61.8  \\

\rowcolor{gray!20}w/o QFS  &69.1  &72.2  &83.8  &73.0  &61.5   \\

\rowcolor{gray!15}w/o QFM  &68.8  &73.9  &83.2  &73.7  &61.3   \\

\rowcolor{gray!10}w/o IFD &68.9  &73.5  &82.8  &73.5  &61.2   \\


\midrule

\end{tabular}
\caption{Ablation study  on NExT-QA and TVQA.}

\label{app:ablation}
\vspace{-7mm}
\end{table}

\begin{figure*}[!tb]
    \centering
    \includegraphics[scale=0.3]{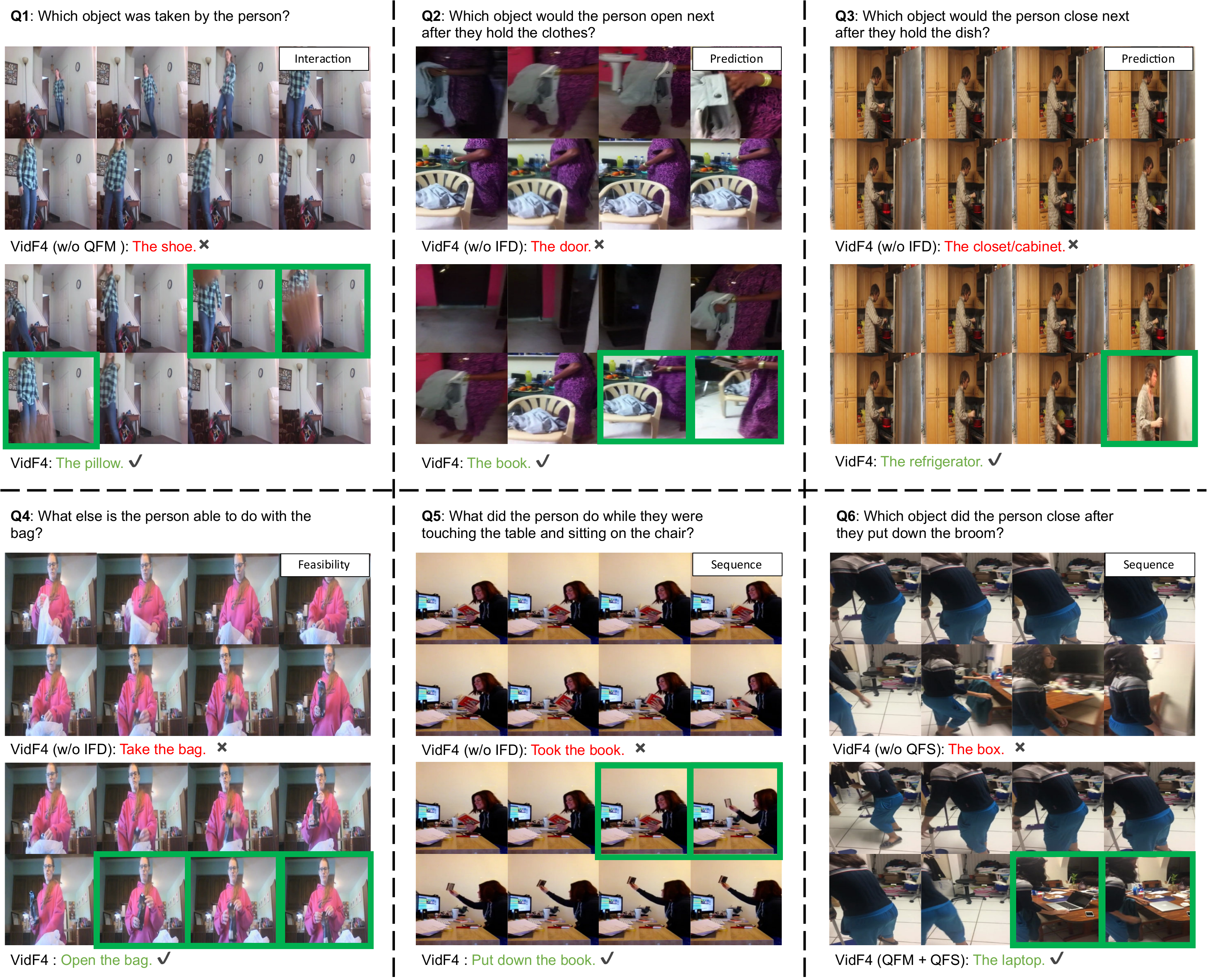}
    \caption{Case study of our \model using 8 Frames. All selected Frames are arranged sequentially from left to right and top to bottom. \model w/o QFM ignores the relevant frames to answer Q1. \model w/o IFD only selects the most relevant but redundant frames, thus unable to answer questions Q2 to Q5. \model successfully identifies the required frames (in \textcolor[rgb]{0,0.7,0.33}{green boxes}) to answer these cases which are sourced from the test set of STAR.}
    \label{fig:example}
    
\end{figure*}

\section{Case Study}
\label{app:cases}
Figure~\ref{fig:example} presents our case study on the Star test set, where we visualize the impact of each component on VidF4's performance. As mentioned in Section~\ref{sec:ifd}, using both QFM and QFS scoring mechanisms may result in high similarity among selected frames. In questions Q2 to Q5 of Figure~\ref{fig:example}, we observe that without IFD, i.e.,  \model (w/o IFD) tends to select frames relevant to the question but with high similarity among them. This is because if a frame has a high QFM and QFS score for a given question, other similar frames are also likely to have high scores, resulting in a group of frames selected by QFM and QFS converging. However, when \model employs IFD, this issue is alleviated to some extent. For instance, in Q2, the IFD mechanism penalizes the scores of frames containing 'white chair' at the same angle, thus encouraging the selection of different frames. Consequently, \model obtains frames showing the action of picking up and holding a 'book' (denoted by the \textcolor[rgb]{0,0.7,0.33}{green boxes} in Q2), enabling it to answer the question correctly. Similar observations can be made in Q3 to Q5 as well.

Furthermore, we note that for \textit{interaction} type questions, the model primarily needs to find frames containing the object mentioned in the question. For example, in Q1, by incorporating QFM, \model locates three frames that include a 'pillow' (as shown in Q1's \textcolor[rgb]{0,0.7,0.33}{green boxes}), allowing it to answer the question correctly. For \textit{Prediction} and \textit{Feasibility} questions, which require the model to possess a certain level of reasoning capability, the model not only needs to provide question-relevant frames but also understand the keywords 'before' or 'after', i.e., frames preceding or following the question-relevant ones. For instance, in Q2 and Q3, the \model must not only identify frames showing the action of 'holding the clothes' or 'holding the dish', but also preserve frames following these actions as much as possible. IFD, by promoting frame diversity, prevents the model from solely focusing on these actions and helps it answer questions more effectively.




\begin{thebibliography}{36}
\providecommand{\natexlab}[1]{#1}

\bibitem[{Chung et~al.(2022)Chung, Hou, Longpre, Zoph, Tay, Fedus, Li, Wang, Dehghani, Brahma et~al.}]{chung2022scaling}
Hyung~Won Chung, Le~Hou, Shayne Longpre, Barret Zoph, Yi~Tay, William Fedus, Yunxuan Li, Xuezhi Wang, Mostafa Dehghani, Siddhartha Brahma, et~al. 2022.
\newblock Scaling instruction-finetuned language models.
\newblock \emph{arXiv preprint arXiv:2210.11416}.

\bibitem[{Efraimidis and Spirakis(2006)}]{efraimidis2006weighted}
Pavlos~S Efraimidis and Paul~G Spirakis. 2006.
\newblock Weighted random sampling with a reservoir.
\newblock \emph{Information processing letters}, 97(5):181--185.

\bibitem[{Fang et~al.(2023)Fang, Wang, Xie, Sun, Wu, Wang, Huang, Wang, and Cao}]{fang2023eva}
Yuxin Fang, Wen Wang, Binhui Xie, Quan Sun, Ledell Wu, Xinggang Wang, Tiejun Huang, Xinlong Wang, and Yue Cao. 2023.
\newblock Eva: Exploring the limits of masked visual representation learning at scale.
\newblock In \emph{Proceedings of the IEEE/CVF Conference on Computer Vision and Pattern Recognition}, pages 19358--19369.

\bibitem[{Gao et~al.(2023)Gao, Zhou, Ji, Zhu, Yang, and Shou}]{gao2023mist}
Difei Gao, Luowei Zhou, Lei Ji, Linchao Zhu, Yi~Yang, and Mike~Zheng Shou. 2023.
\newblock Mist: Multi-modal iterative spatial-temporal transformer for long-form video question answering.
\newblock In \emph{Proceedings of the IEEE/CVF Conference on Computer Vision and Pattern Recognition}, pages 14773--14783.

\bibitem[{Hannane et~al.(2016)Hannane, Elboushaki, Afdel, Naghabhushan, and Javed}]{hannane2016efficient}
Rachida Hannane, Abdessamad Elboushaki, Karim Afdel, P~Naghabhushan, and Mohammed Javed. 2016.
\newblock An efficient method for video shot boundary detection and keyframe extraction using sift-point distribution histogram.
\newblock \emph{International Journal of Multimedia Information Retrieval}, 5:89--104.

\bibitem[{Huang and Wang(2019)}]{huang2019novel}
Cheng Huang and Hongmei Wang. 2019.
\newblock A novel key-frames selection framework for comprehensive video summarization.
\newblock \emph{IEEE Transactions on Circuits and Systems for Video Technology}, 30(2):577--589.

\bibitem[{Jang et~al.(2017)Jang, Song, Yu, Kim, and Kim}]{jang2017tgif}
Yunseok Jang, Yale Song, Youngjae Yu, Youngjin Kim, and Gunhee Kim. 2017.
\newblock Tgif-qa: Toward spatio-temporal reasoning in visual question answering.
\newblock In \emph{Proceedings of the IEEE conference on computer vision and pattern recognition}, pages 2758--2766.

\bibitem[{Khan et~al.(2020)Khan, Mazaheri, Lobo, and Shah}]{khan2020mmft}
Aisha~Urooj Khan, Amir Mazaheri, Niels Da~Vitoria Lobo, and Mubarak Shah. 2020.
\newblock Mmft-bert: Multimodal fusion transformer with bert encodings for visual question answering.
\newblock \emph{arXiv preprint arXiv:2010.14095}.

\bibitem[{Kim et~al.(2020)Kim, Tang, and Bansal}]{kim2020dense}
Hyounghun Kim, Zineng Tang, and Mohit Bansal. 2020.
\newblock Dense-caption matching and frame-selection gating for temporal localization in videoqa.
\newblock In \emph{Proceedings of the 58th Annual Meeting of the Association for Computational Linguistics}, pages 4812--4822.

\bibitem[{Kim et~al.(2021)Kim, Jeong, Kim, Kang, and Kwak}]{kim2021self}
Seonhoon Kim, Seohyeong Jeong, Eunbyul Kim, Inho Kang, and Nojun Kwak. 2021.
\newblock Self-supervised pre-training and contrastive representation learning for multiple-choice video qa.
\newblock In \emph{Proceedings of the AAAI Conference on Artificial Intelligence}, volume~35, pages 13171--13179.

\bibitem[{Ko et~al.(2023)Ko, Lee, Kang, Roh, and Kim}]{ko2023large}
Dohwan Ko, Ji~Lee, Woo-Young Kang, Byungseok Roh, and Hyunwoo Kim. 2023.
\newblock Large language models are temporal and causal reasoners for video question answering.
\newblock In \emph{Proceedings of the 2023 Conference on Empirical Methods in Natural Language Processing}, pages 4300--4316.

\bibitem[{Kulhare et~al.(2016)Kulhare, Sah, Pillai, and Ptucha}]{kulhare2016key}
Sourabh Kulhare, Shagan Sah, Suhas Pillai, and Raymond Ptucha. 2016.
\newblock Key frame extraction for salient activity recognition.
\newblock In \emph{2016 23rd International Conference on Pattern Recognition (ICPR)}, pages 835--840. IEEE.

\bibitem[{Lei et~al.(2021)Lei, Li, Zhou, Gan, Berg, Bansal, and Liu}]{lei2021less}
Jie Lei, Linjie Li, Luowei Zhou, Zhe Gan, Tamara~L Berg, Mohit Bansal, and Jingjing Liu. 2021.
\newblock Less is more: Clipbert for video-and-language learning via sparse sampling.
\newblock In \emph{Proceedings of the IEEE/CVF conference on computer vision and pattern recognition}, pages 7331--7341.

\bibitem[{Lei et~al.(2018)Lei, Yu, Bansal, and Berg}]{lei2018tvqa}
Jie Lei, Licheng Yu, Mohit Bansal, and Tamara~L Berg. 2018.
\newblock Tvqa: Localized, compositional video question answering.
\newblock \emph{arXiv preprint arXiv:1809.01696}.

\bibitem[{Li et~al.(2023)Li, Li, Savarese, and Hoi}]{li2023blip}
Junnan Li, Dongxu Li, Silvio Savarese, and Steven Hoi. 2023.
\newblock Blip-2: Bootstrapping language-image pre-training with frozen image encoders and large language models.
\newblock \emph{arXiv preprint arXiv:2301.12597}.

\bibitem[{Pan et~al.(2023)Pan, Lin, Ge, Zhu, Zhang, Wang, Qiao, and Li}]{pan2023retrieving}
Junting Pan, Ziyi Lin, Yuying Ge, Xiatian Zhu, Renrui Zhang, Yi~Wang, Yu~Qiao, and Hongsheng Li. 2023.
\newblock Retrieving-to-answer: Zero-shot video question answering with frozen large language models.
\newblock \emph{arXiv preprint arXiv:2306.11732}.

\bibitem[{Patel et~al.(2021)Patel, Parikh, and Shastri}]{patel2021recent}
Devshree Patel, Ratnam Parikh, and Yesha Shastri. 2021.
\newblock Recent advances in video question answering: A review of datasets and methods.
\newblock In \emph{Pattern Recognition. ICPR International Workshops and Challenges: Virtual Event, January 10--15, 2021, Proceedings, Part II}, pages 339--356. Springer.

\bibitem[{Singh et~al.(2021)Singh, Meetei, Singh, Singh, and Bandyopadhyay}]{singh2021efficient}
Alok Singh, Loitongbam~Sanayai Meetei, Salam~Michael Singh, Thoudam~Doren Singh, and Sivaji Bandyopadhyay. 2021.
\newblock An efficient keyframes selection based framework for video captioning.
\newblock In \emph{Proceedings of the 18th International Conference on Natural Language Processing (ICON)}, pages 240--250.

\bibitem[{Tang et~al.(2019)Tang, Liu, Xiao, and Sebe}]{tang2019fast}
Hao Tang, Hong Liu, Wei Xiao, and Nicu Sebe. 2019.
\newblock Fast and robust dynamic hand gesture recognition via key frames extraction and feature fusion.
\newblock \emph{Neurocomputing}, 331:424--433.

\bibitem[{Tapaswi et~al.(2016)Tapaswi, Zhu, Stiefelhagen, Torralba, Urtasun, and Fidler}]{tapaswi2016movieqa}
Makarand Tapaswi, Yukun Zhu, Rainer Stiefelhagen, Antonio Torralba, Raquel Urtasun, and Sanja Fidler. 2016.
\newblock Movieqa: Understanding stories in movies through question-answering.
\newblock In \emph{Proceedings of the IEEE conference on computer vision and pattern recognition}, pages 4631--4640.

\bibitem[{Touvron et~al.(2023)Touvron, Lavril, Izacard, Martinet, Lachaux, Lacroix, Rozi{\`e}re, Goyal, Hambro, Azhar et~al.}]{touvron2023llama}
Hugo Touvron, Thibaut Lavril, Gautier Izacard, Xavier Martinet, Marie-Anne Lachaux, Timoth{\'e}e Lacroix, Baptiste Rozi{\`e}re, Naman Goyal, Eric Hambro, Faisal Azhar, et~al. 2023.
\newblock Llama: Open and efficient foundation language models.
\newblock \emph{arXiv preprint arXiv:2302.13971}.

\bibitem[{Wang et~al.(2023)Wang, Ge, Yan, Ge, Lin, Tsutsui, Lin, Cai, Wu, Shan et~al.}]{wang2023all}
Jinpeng Wang, Yixiao Ge, Rui Yan, Yuying Ge, Kevin~Qinghong Lin, Satoshi Tsutsui, Xudong Lin, Guanyu Cai, Jianping Wu, Ying Shan, et~al. 2023.
\newblock All in one: Exploring unified video-language pre-training.
\newblock In \emph{Proceedings of the IEEE/CVF Conference on Computer Vision and Pattern Recognition}, pages 6598--6608.

\bibitem[{Wang et~al.(2022)Wang, Li, Li, He, Huang, Zhao, Zhang, Xu, Liu, Wang et~al.}]{wang2022internvideo}
Yi~Wang, Kunchang Li, Yizhuo Li, Yinan He, Bingkun Huang, Zhiyu Zhao, Hongjie Zhang, Jilan Xu, Yi~Liu, Zun Wang, et~al. 2022.
\newblock Internvideo: General video foundation models via generative and discriminative learning.
\newblock \emph{arXiv preprint arXiv:2212.03191}.

\bibitem[{Wang et~al.(2024)Wang, Wang, Wu, Liang, Zhao, and Zheng}]{wang2024lstp}
Yuxuan Wang, Yueqian Wang, Pengfei Wu, Jianxin Liang, Dongyan Zhao, and Zilong Zheng. 2024.
\newblock Lstp: Language-guided spatial-temporal prompt learning for long-form video-text understanding.
\newblock \emph{arXiv preprint arXiv:2402.16050}.

\bibitem[{Wu et~al.(2021)Wu, Yu, Chen, Tenenbaum, and Gan}]{wu2021star}
Bo~Wu, Shoubin Yu, Zhenfang Chen, Joshua~B Tenenbaum, and Chuang Gan. 2021.
\newblock Star: A benchmark for situated reasoning in real-world videos.
\newblock In \emph{Thirty-fifth conference on neural information processing systems datasets and benchmarks track (Round 2)}.

\bibitem[{Xiao et~al.(2021)Xiao, Shang, Yao, and Chua}]{xiao2021next}
Junbin Xiao, Xindi Shang, Angela Yao, and Tat-Seng Chua. 2021.
\newblock Next-qa: Next phase of question-answering to explaining temporal actions.
\newblock In \emph{Proceedings of the IEEE/CVF conference on computer vision and pattern recognition}, pages 9777--9786.

\bibitem[{Xie and Ermon(2019)}]{xie2019reparameterizable}
Sang~Michael Xie and Stefano Ermon. 2019.
\newblock Reparameterizable subset sampling via continuous relaxations.
\newblock \emph{arXiv preprint arXiv:1901.10517}.

\bibitem[{Xu et~al.(2017)Xu, Zhao, Xiao, Wu, Zhang, He, and Zhuang}]{xu2017video}
Dejing Xu, Zhou Zhao, Jun Xiao, Fei Wu, Hanwang Zhang, Xiangnan He, and Yueting Zhuang. 2017.
\newblock Video question answering via gradually refined attention over appearance and motion.
\newblock In \emph{Proceedings of the 25th ACM international conference on Multimedia}, pages 1645--1653.

\bibitem[{Yan et~al.(2020)Yan, Gilani, Feng, Zhang, Qin, and Mian}]{yan2020self}
Xiang Yan, Syed~Zulqarnain Gilani, Mingtao Feng, Liang Zhang, Hanlin Qin, and Ajmal Mian. 2020.
\newblock Self-supervised learning to detect key frames in videos.
\newblock \emph{Sensors}, 20(23):6941.

\bibitem[{Yang et~al.(2021)Yang, Miech, Sivic, Laptev, and Schmid}]{yang2021just}
Antoine Yang, Antoine Miech, Josef Sivic, Ivan Laptev, and Cordelia Schmid. 2021.
\newblock Just ask: Learning to answer questions from millions of narrated videos.
\newblock In \emph{Proceedings of the IEEE/CVF international conference on computer vision}, pages 1686--1697.

\bibitem[{Yang et~al.(2020)Yang, Garcia, Chu, Otani, Nakashima, and Takemura}]{yang2020bert}
Zekun Yang, Noa Garcia, Chenhui Chu, Mayu Otani, Yuta Nakashima, and Haruo Takemura. 2020.
\newblock Bert representations for video question answering.
\newblock In \emph{Proceedings of the IEEE/CVF Winter Conference on Applications of Computer Vision}, pages 1556--1565.

\bibitem[{Ye et~al.(2023)Ye, Xu, Yan, Xu, Qian, Zhang, and Huang}]{ye2023hitea}
Qinghao Ye, Guohai Xu, Ming Yan, Haiyang Xu, Qi~Qian, Ji~Zhang, and Fei Huang. 2023.
\newblock Hitea: Hierarchical temporal-aware video-language pre-training.
\newblock In \emph{Proceedings of the IEEE/CVF International Conference on Computer Vision}, pages 15405--15416.

\bibitem[{Yu et~al.(2023)Yu, Cho, Yadav, and Bansal}]{yu2023self}
Shoubin Yu, Jaemin Cho, Prateek Yadav, and Mohit Bansal. 2023.
\newblock Self-chained image-language model for video localization and question answering.
\newblock \emph{arXiv preprint arXiv:2305.06988}.

\bibitem[{Zhang et~al.(2023)Zhang, Li, and Bing}]{zhang2023video}
Hang Zhang, Xin Li, and Lidong Bing. 2023.
\newblock Video-llama: An instruction-tuned audio-visual language model for video understanding.
\newblock \emph{arXiv preprint arXiv:2306.02858}.

\bibitem[{Zhao et~al.(2018)Zhao, Zhang, Xiao, Yu, Yu, Cai, Wu, and Zhuang}]{zhao2018open}
Zhou Zhao, Zhu Zhang, Shuwen Xiao, Zhou Yu, Jun Yu, Deng Cai, Fei Wu, and Yueting Zhuang. 2018.
\newblock Open-ended long-form video question answering via adaptive hierarchical reinforced networks.
\newblock In \emph{IJCAI}, volume~2, page~8.

\bibitem[{Zhong et~al.(2022)Zhong, Xiao, Ji, Li, Deng, and Chua}]{zhong2022Video}
Yaoyao Zhong, Junbin Xiao, Wei Ji, Yicong Li, Weihong Deng, and Tat-Seng Chua. 2022.
\newblock Video question answering: Datasets, algorithms and challenges.
\newblock In \emph{The 2022 Conference on Empirical Methods in Natural Language Processing}.

\end{thebibliography}
\end{document}